\documentclass[
twocolumn,
]{ceurart}

\usepackage{siunitx}

\usepackage{listings}

\begin{document}

\copyrightyear{2022}
\copyrightclause{Copyright for this paper by its authors.
  Use permitted under Creative Commons License Attribution 4.0
  International (CC BY 4.0).}

\conference{RecSys in HR'22: The 2nd Workshop on Recommender Systems for Human Resources, in conjunction with the 16th ACM Conference on Recommender Systems, September 18--23, 2022, Seattle, USA.}

\title{Closing the Gender Wage Gap: Adversarial Fairness in Job Recommendation}

\author[1]{Clara Rus}[%
email=clara.rus@ru.nl,
]

\address[1]{Radboud University, Houtlaan 4, 6525XZ, Nijmegen, the Netherlands}
\address[2]{DPG Media Online Services, Jacob Bontiusplaats 9, 1018LL, Amsterdam, the Netherlands}

\author[2]{Jeffrey Luppes}[%
orcid=0000-0002-0465-535X,
email=jeffrey.luppes@dpgmedia.nl,
]

\author[1]{Harrie Oosterhuis}[
orcid= 0000-0002-0458-9233,
email=harrie.oosterhuis@ru.nl,
]

\author[2]{Gido H. Schoenmacker}[%
orcid=0000-0003-3946-928X,
email=gido.schoenmacker@dpgmedia.nl,
]

\begin{abstract}
The goal of this work is to help mitigate the already existing gender wage gap by supplying unbiased job recommendations based on resumes from job seekers. We employ a generative adversarial network to remove gender bias from word2vec representations of 12M job vacancy texts and 900k resumes. Our results show that representations created from recruitment texts contain algorithmic bias and that this bias results in real-world consequences for recommendation systems. Without controlling for bias, women are recommended jobs with significantly lower salary in our data. With adversarially fair representations, this wage gap disappears, meaning that our debiased job recommendations reduce wage discrimination. We conclude that adversarial debiasing of word representations can increase real-world fairness of systems and thus may be part of the solution for creating fairness-aware recommendation systems. 
\end{abstract}


\maketitle

\section{Introduction}

The recruitment industry relies more and more on automation for processing, searching, and matching job vacancies to job seekers. However, automation of the recruitment process can lead to discriminatory results with respect to certain groups, based on gender, ethnicity or age~\cite{kochling2020discriminated}. Inequality in employment and remuneration still exists between for example ethnic groups~\cite{thijssen2021discrimination, bisschop2020ethnic, ramos2021labour} and gender groups~\cite{matteazzi2018part, ciminelli2021sticky}, thus naive implementations of AI recruitment systems are at risk of copying and perpetuating these inequalities. 

One reason for an algorithm to show discriminatory behaviour is the input data~\cite{chouldechova2020snapshot}. If the data is under--representative or if historical bias is present, then the system can propagate this in its predictions~\cite{kochling2020discriminated}. Ignoring the presence of bias in the data, can perpetuate existing (gender) stereotypes and inequalities in employment. 

Examples of systems that have shown biased behaviour with respect to gender include the Amazon recruitment system\footnote{\href{https://www.reuters.com/article/idUSKCN1MK08G}{https://www.reuters.com/article/idUSKCN1MK08G}} and the Facebook Add algorithm~\cite{ali2019discrimination}. Also widely used models, such as BERT~\cite{devlin2018bert} and word2vec~\cite{mikolov2013efficient}, have been shown to create biased representations~\cite{kurita2019measuring, bolukbasi2016man}. Obtaining fair representations could eliminate the bias present in the data and help a system achieve fairer predictions~\cite{https://doi.org/10.48550/arxiv.1707.00075}.

One way to learn debiased representations is through adversarial learning. State-of-the-art adversarial debiasing methods~\cite{edwards2015censoring,xu2019fairgan+,madras2018learning,wu2021fairness,liu2022dual} rely on the same general approach as generative adversarial networks~\cite{https://doi.org/10.48550/arxiv.1406.2661}. A generator model is trained to produce new data representations, that are critiqued by an adversary neural network. The adversary tries to predict the sensitive variable (in our case, gender) from the produced representation. By training the representations together with an adversary and classifier, they are aimed to be both fair and useful for the task.

This work is motivated by the desire to supply unbiased job recommendations to job seekers. We focus specifically on mitigating gender bias in word embeddings obtained from recruitment texts using adversarial learning. Our work adds to existing research by applying state-of-the-art debiasing~\cite{edwards2015censoring,zdenizci2020} to industrial sized free-format recruitment textual data. Firstly, we investigate gender bias in the existing representations and the unfairness it results in. Secondly, we apply two debiasing methods to create new representations. These methods balance multi-label classification to ensure that task-relevant information has been preserved, with an adversarial setup that attempts to remove the effects of gender bias. The resulting new representations are tested in a job recommendation setting where the difference in wage between jobs recommended based on female/male resumes is evaluated.

To summarize, our contributions are three-fold: (i) we measure whether adversarial learning can mitigate gender bias in representations of industrial sized free-format recruitment textual data; (ii) we show whether debiased representations help achieve fairness and performance on a multi-label classification task; and (iii) to the authors' best knowledge, we are the first to successfully apply debiased representations to help solve the gender wage-gap in a job recommendation setting. Moreover, our implementation of the adversarial debiasing method is publicly available.

In the next section, our data and methods are described in detail. After that, the results are presented. Lastly, these results are discussed together with our final conclusions and suggestions for future directions.

\section{Data and Method}

\subsection{Data}
The recruitment data set used throughout this research consists of job vacancies and job seeker information provided by DPG Recruitment. Job vacancy information included (i) salary ranges, (ii) working hours, and (iii) anonymised free-format job vacancy texts. In total there are 12 millions vacancies.

Job seeker information consisted of (i) one or more industry group(s) that the job seeker expressed interest in (out of a total of 21 pre-defined groups), (ii) inferred dichotomous gender, and (iii) anonymised free-format resume texts. Gender of the job seeker was inferred based on first name. From the total of available resumes, entries with missing data (65\%) or ambiguous first name (3\%) were excluded, leaving 904,576 (32\%) complete resumes with a female to male ratio of 0.93. Anonymisation included removal of all names (including company names), dates, addresses, telephone numbers, email addresses, websites, and other contact information. A more complete overview of this data is given in Appendix~\ref{app:resumes}.

Both vacancy and resume texts were embedding into 300-dimensional word vector using a word2vec~\cite{mikolov2013efficient} model trained on all vacancy texts. Finally, each text was represented as the mean over the embeddings of the words composing the text.

\subsection{Bias and debiasing}

Previous research has shown that popular models such as BERT~\cite{devlin2018bert} and word2vec~\cite{mikolov2013efficient} can create biased representations~\cite{kurita2019measuring, bolukbasi2016man,WEAT}. In this work, two debiasing methods were employed to combat this bias.

Firstly, to create a simple baseline, we attempt to debias the representations by replacing gendered words with neutral words. For example, gendered pronouns ``she''/``he'', ``her''/``his'' are replaced with neutral pronouns ``they'' and ``theirs''. Gendered words such as: ``woman''/``man'', ``girl''/``boy'' are replaced with the word ``person''. The full list of substitutions can be found in Appendix~\ref{app:words}. A new word2vec model was trained on this augmented corpus, resulting in new representations for both the resumes and the vacancies. In the remaining text, ``original representations'' will refer to the representations trained on the original texts, whereas ``word-substitution representations'' will refer to the representations trained on the altered texts.

\begin{figure}
\begin{center}
\includegraphics[width=0.48\textwidth]{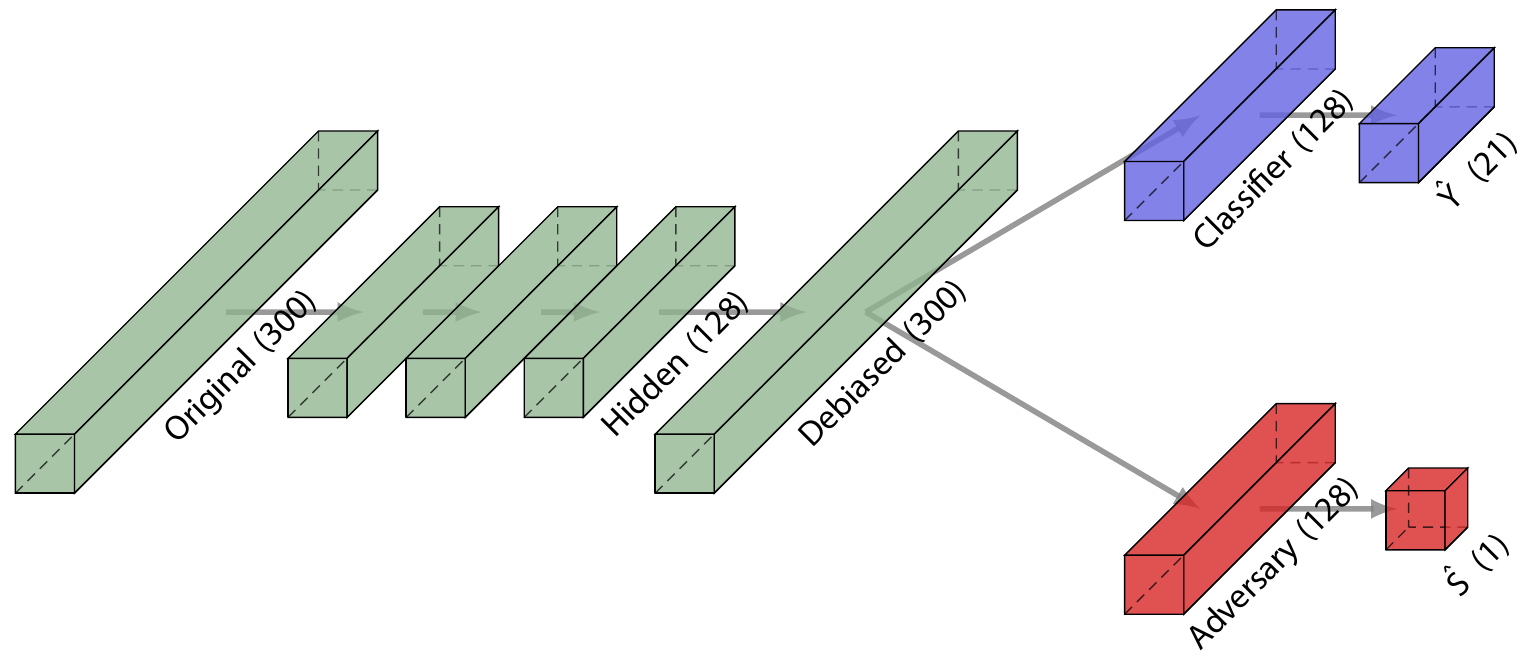}
\end{center}
\caption{Architecture of the adversarial setup. The left section (green) represents the generator, consisting of an input layer (d=300) for the word2vec representations, three hidden layers (d=128), and an output layer (d=300) for the debiased representations. The top section (blue) represents the classifier consisting of a hidden layer (d=128) and an output layer $\hat{Y}$ (d=21) encoding the industry groups. The bottom section (red) represents the adversary and consists of a hidden layer (d=128) and an output neuron $\hat{S}$ (d=1) encoding the sensitive variable gender.}
\label{fig:nn}
\end{figure}

Secondly, we applied the adversarial approach as proposed by \citet{edwards2015censoring}. This method consists of three neural network components: a generator, a classifier, and an adversary. Inspired by \citet{zdenizci2020}, we chose the following the architecture: The generator is a multilayer perceptron with three hidden layers of $128$ neurons that outputs a $300$ dimensional vector representing the new representations. The classifier and the adversary have one hidden layer of $128$ neurons. The output dimension of the classifier is $21$ (industry group classes), and the output dimension of the adversary is one (gender). An architecture schematic is included as Figure \ref{fig:nn}.

The generator creates new representations for the classification task, while the adversary attempts to predict a sensitive variable gender from these new representations. The goal of the generator is to create representations that can fool the adversary in such a way that the sensitive variable can no longer be predicted, while also obtaining a good performance on the classification task. The classification task is considered to be a multi-label task of 21 classes, predicting the industry group(s) for each job seeker. This means that the classification loss should be minimized while the adversarial loss should be maximized. The final loss (Equation \ref{eq:1}) of the model is a weighted sum of the adversarial loss and the classification loss, where Z are the newly generated representations, Y' are the predictions of the classifier and S' are the predictions of the adversary:
\begin{equation}
L = \alpha L_{cla}(Z,Y') + \beta L_{adv}(Z,S').
\label{eq:1}
\end{equation}
We will call representations created by this method  ``adversarial representations''. Because the adversarial process could be unstable, all results pertaining to these are the mean of $5$ independent complete training runs.

\subsection{Evaluation}
 
Classifiers for both industry groups and sensitive variable are evaluated in terms of accuracy and area under the receiver operating characteristic curve (AUC). Fairness was evaluated using statistical parity~\cite{mehrabi2021survey}:

 \begin{equation}
 P(cla(Z)=1|S = 1) - P(cla(Z)=1|S = 0) < \epsilon. \label{eq:2}
 \end{equation}

In the recruitment industry, if a system designed to match resumes and vacancies perpetuates biased associations, it could lead to a wage gap between salaries of women and men~\cite{calanca2019responsible}. To specifically test differences in salary, a salary association test was performed between the representations of the resumes and of the vacancies. Using the embeddings of the resumes and the vacancies the L2 distance matrix was computed and each resume was matched to the closest vacancy. The salary distribution of the matched vacancies of the female-inferred group were compared with the male-inferred group.
 
\subsection{Experimental Setup}

The train split was created by taking 30\% of random samples for the validation split, and the rest of the full data is used for training. The full data set was not used for the salary association due to computational limitations. Instead, 10,000 resumes were associated with all jobs from the time period June 2020--June 2021 that provided salary information. This resulted in 23,501 total vacancies. All experiments were conducted using a fixed 70-30\% split and the Adam optimizer with a learning rate of $1\mathrm{e}{-5}$. For all components the binary cross-entropy loss was used. Parameters of the final loss (Equation \ref{eq:1}) are set in the following way: $\alpha$ = 1, $\beta$ = 1. The implementation of the adversarial debiasing method can be found at: \url{https://github.com/ClaraRus/Debias-Embeddings-Recruitment}.

\section{Results}

\subsection{Prediction of sensitive variable}

\noindent Firstly, the discriminatory power to predict the sensitive variable gender was tested using the original, word-substitution, and adversarial representations. Using the original representations, gender is predicted with 94\% AUC and an accuracy of 86\%; the word-substitution representations result in an 93\% AUC and an accuracy of 85\%; lastly, the adversarial representations lowered both the accuracy and the AUC to 82\%.

\subsection{Prediction of industry group}

\noindent Secondly, the information contents and statistical parity of the three representation types were tested by attempting to predict the function group based on resume representation. Table \ref{tab:fairness_performance} shows the result obtained in terms of performance and statistical parity. 

Training a classifier with the original representations of the resumes obtained a statistical parity of $0.076$. The word-substitution representations obtained similar results. Using an adversarial approach improved the statistical parity by $21$\%, at the cost of lowering the accuracy by $2$ percentage point and the true positive rate by $16$ percentage point. 

\begin{table*}[!htbp]
\caption{Statistical parity and performance of multi-label classification of 21 industry groups from three different types of representations. Original representations were obtained using word2vec. Word-substitution representations were obtained using a word-substitution debiasing method. Adversarial representations were obtained using the adversarial debiasing method. The ``Overall'' row represents the weighted mean. Parity: statistical parity (Eq. \ref{eq:2}), closer to zero is better. TPR: True positive rate.} 
\centering
\resizebox{\textwidth}{!}{
\begin{tabular}{l|ccc|ccc|ccc}
\toprule
 & \multicolumn{3}{c}{\textbf{Original}} & \multicolumn{3}{c}{\textbf{Word-substitution}} & \multicolumn{3}{c}{\textbf{Adversarial}}\\
  & \hspace{0cm} Parity & \hspace{0cm} Accuracy & \hspace{0cm} TPR  & \hspace{0cm} Parity & \hspace{0cm} Accuracy & \hspace{0cm} TPR & \hspace{0cm} Parity & \hspace{0cm} Accuracy & \hspace{0cm} TPR  \\
\midrule
Overall & 0.076 & 0.90 & 0.37 & 0.760 & 0.90 & 0.38 & 0.060 & 0.89 & 0.21 \\
    Administration/Secretarial &  0.267 &  0.74 &  0.52 & 0.271 & 0.74 &  0.53 & 0.260 & 0.72 & 0.45 \\
    Automation/Internet & 0.066 & 0.83 & 0.46 & 0.069 & 0.83 &  0.47 & 0.045 & 0.82 & 0.32   \\
    Policy/Executive & 0.000 &   0.79 &   0.19 & 0.001 & 0.79 &  0.21 & 0.005 & 0.79 & 0.08 \\
    Security/Defence/Police & 0.010 &   0.83 &   0.16 & 0.009 & 0.83 &  0.16 & 0.000 &  0.83 & 0.03 \\
    Commercial/Sales & 0.074 &   0.74 &   0.36 & 0.070 & 0.74 &  0.35 & 0.059 &  0.72 &  0.19 \\
    Consultancy/Advice &  0.026 &   0.75 &   0.20 & 0.033 & 0.75 &  0.22 & 0.012 &  0.74 & 0.07 \\
    Design/Creative/Journalism & 0.004 &   0.82 &   0.26 & 0.005 & 0.82 &  0.30 & 0.001 & 0.82 & 0.09 \\
    Management & 0.070 &   0.78 &   0.32 & 0.063 &  0.78 &  0.29 &  0.052 & 0.77 & 0.22   \\
    Financial/Accounting & 0.021 &   0.81 &   0.47 & 0.021 & 0.81 &  0.48 &  0.020 & 0.80 & 0.29  \\ 
   Financial services &  0.012 &   0.79 &   0.22 & 0.012 & 0.79 &  0.28 & 0.007 & 0.79 & 0.10  \\
    HR/Training & 0.041 &   0.80 &   0.32 & 0.043 &  0.80 &  0.34 & 0.014 & 0.78 & 0.09  \\
    Catering/Retail &  0.037 &   0.76 &   0.33 & 0.023 & 0.76 &  0.27 & 0.018 & 0.75 & 0.14 \\
    Procurement/Logistics/Transport &  0.115 &   0.77 &   0.38 & 0.102 & 0.77 &  0.35 & 0.087 &  0.76 & 0.24 \\
    Legal & 0.015 &   0.85 &   0.45 & 0.015 &  0.85 &  0.44 & 0.002 &  0.84 &     0.09  \\
    Customer service/Call centre/Front office & 0.039 &   0.76 &   0.12 & 0.031 & 0.76 &  0.10 & 0.001 & 0.76 & 0.01 \\
    Marketing/PR/Communications & 0.031 &   0.77 &   0.41 & 0.031 & 0.77 &  0.45 & 0.028 &     0.76 &     0.27 \\
    Medical/Healthcare & 0.115 &   0.76 &   0.40 & 0.116 & 0.77 &  0.40  & 0.100 &     0.75 &     0.27 \\
    Education/Research/Science & 0.045 &   0.77 &   0.32 & 0.057 &      0.77 &  0.39 & 0.031 &     0.75 &     0.16 \\
    Other & 0.005 &   0.68 &   0.04 & 0.009 &      0.68 &  0.05 & 0.000 &   0.68 &     0.00 \\
    Production/Operational & 0.063 &   0.78 &   0.27 & 0.062 &  0.78 &  0.27 & 0.043 &  0.77 &     0.15 \\
    Technology & 0.165 &   0.79 &   0.51 & 0.169 &  0.79 &  0.52 & 0.153 &  0.78 &  0.43 \\
\bottomrule
\end{tabular}}
\label{tab:fairness_performance}
\end{table*}

\begin{table*}[!htbp]
\caption{Salary Association Test between resumes and vacancies. For each resume the most similar vacancy was assigned based on Euclidean distance in the representation space. The values represent the salary per hour in Euros (\texteuro). Original representations were obtained using word2vec. Word-substitution representations were obtained using the word-substitution debiasing method. Adversarial representations were obtained using the adversarial debiasing method. The top three rows represent the weighted summary statistics. The industries names with an asterisk (*) are the ones for which the adversarial method reduced the wage gap.} 
\centering
\resizebox{\textwidth}{!}{
\begin{tabular}{l|ccc|ccc|ccc}
\toprule
 & \multicolumn{3}{c}{\textbf{Original}} & \multicolumn{3}{c}{\textbf{Word-substitution}} & \multicolumn{3}{c}{\textbf{Adversarial}}\\

 &    Female &     Male &  Wage gap &   Female &     Male &  Wage gap &   Female &     Male & Wage gap  \\
\midrule
Mean & 25.28 & 26.09 & 0.81 & 25.19 & 26.14 & 0.95 & 27.06 & 27.15 & 0.09 \\
Standard deviation & 9.43 & 9.90 & 0.47 & 9.54 & 10.07 & 0.53 & 10.14 & 9.94 & -0.20 \\
Median & 23.40 & 23.62 & 0.22 & 22.95 & 23.62 & 0.67 & 23.97 & 24.30 & 0.33\\
\cmidrule(lr{.5em}){1-1}
Administration/Secretarial* & 
23.50 &   24.94 &   1.44 &   23.44 &   24.95 &  1.51 &   26.45 &    26.41 & -0.04   \\
Automation/Internet &      
   28.34 &   28.58 &  0.24 &  28.05 &   29.02 &   0.97 & 29.94 &    28.44 & -1.50   \\
Policy/Executive* &   29.90 &   31.23 &   1.33 &   30.16 &   31.53 &   1.37 &   30.35 &    31.18 & 0.83  \\
Security/Defence/Police* &    
24.81 &   22.78 &   -2.03 &   24.81 &   23.09 & -1.72 &    25.51 &    26.44 & 0.93  \\
Commercial/Sales* &  23.76 &   25.77 &  2.01 &  23.88 &   25.53 &  1.65 & 26.66 &    27.39 & 0.73   \\
Consultancy/Advice* &   29.27 &   30.49 &  1.22 &  29.25 &   30.42 &  1.17 &   29.92 &    30.63 &  0.71 \\
Design/Creative/Journalism &  26.39 &   26.33 &  -0.06 &   26.13 &   26.12 &  -0.01 &    28.22 &    28.02 &  -0.20  \\
Management* &    29.49 &   31.22 & 1.73 &    29.49 &   31.47 &   1.98 &   30.66 &    30.31 &  -0.35  \\
Financial/Accounting* &   24.30 &   27.94 &  3.64 &   24.43 &   28.07 &   3.64 &    27.20 &    28.62 & 1.42  \\
Financial services* & 24.33 &   27.85 &  3.52 &  24.19 &   27.76 &   3.57 & 26.80 &    28.67 & 1.87\\
HR/Training &  
28.59 &   28.87 &  0.28 &    28.80 &   29.10 &    0.30 &      29.52 &    29.15 &  -0.37  \\
Catering/Retail* &  22.80 &   23.76 &  0.96 &  22.76 &   23.49 &  0.73 &25.15 &   24.50 & 0.65  \\
Procurement/Logistics/Transport & 
23.70 &   23.28 &  -0.42 &  23.46 &   23.30 &  -0.16 &  25.96 &    25.10  & -0.86  \\
Legal* & 
24.89 &   28.79 & 3.90 &25.52 &   28.91 &  3.39 & 28.82 &    29.01 & 0.19    \\
Customer service/Call centre/Front office* &   22.89 &   23.78 &   0.89 &    23.01 &   23.85 &    0.84 &    25.35 &    25.96 &  0.61   \\
Marketing/PR/Communications* & 
26.64 &   27.65 &   1.01 &   26.71 &   27.55 &   0.84 & 28.86 &    29.22 &  0.36  \\
Medical/Healthcare* & 
26.30 &   27.51 &   1.21 &  26.11 &   27.19 &   1.08 &  27.17 &    28.07 & 0.90      \\
Education/Research/Science & 
28.82 &   27.43 &   -1.39 & 28.30 &   27.65 &   -0.65 & 27.66 &   29.07 &   1.41   \\
Other* &    
24.91 &   24.58 &  -0.33 &  24.79 &   24.84 & 0.05 & 26.07 &   26.32 &    0.25  \\
Production/Operational* & 
21.69 &   22.61 & 0.92 
&  21.15 &   22.49 & 1.34  & 24.39 &    23.94 & -0.45\\
Technology* &       25.51 &   24.09 &  -1.42 &  24.57 &   24.07 & -0.50 &   25.79 &    25.79 & 0.00 \\
\bottomrule
\end{tabular}}
\label{tab:salary_test_eng_nl_industries}
\end{table*}

\subsection{Salary Association Test}

Thirdly, a salary association test was performed using the three representation types. Table~\ref{tab:salary_test_eng_nl_industries} describes the salary distribution of the female and male groups for each debiasing method. In the female group there are 4827 samples and in the male group 5173.

Using the original representations, female-inferred resumes were associated with a mean salary of \texteuro25.28 per hour, whereas male-inferred resumes were associated with a mean salary of \texteuro26.09 per hour, which is significantly (p<$1\mathrm{e}{-5}$) higher. This results in an estimated average annual wage gap of \texteuro1680.

Using the word-substitution representations, female-inferred resumes were associated with a mean salary of \texteuro25.19 per hour, whereas male-inferred resumes were associated with a mean salary of \texteuro26.14 per hour. The difference between the means of the female group and the male group increased, broadening the annual wage gap to \texteuro1900 (with a significant difference between groups, p<$1\mathrm{e}{-7}$).

With the adversarial representations, female-inferred resumes were associated with a mean salary of \texteuro27.06 an hour, whereas male-inferred resumes were associated with a mean salary of \texteuro27.15 an hour. Using the adversarial method to generate fair representations for both the resumes and vacancies decreased the mean gap, lowering the annual wage gap to \texteuro180. This resulted in the female/male difference now being non-significant (p=$0.47$).

Table \ref{tab:salary_test_eng_nl_industries} shows the mean salary per hour for each industry group. Ideally females and males belonging to the same industry group should have similar salaries. The word-substitution representations lowered the wage gap in 13 of the industry groups by \texteuro620 per year, while increasing the gap for 7 with an average of \texteuro460 per year. For ``Financial/Accounting'' there is no change in the salary association. The adversarial method lowered the wage gap in 16 out of the 21 industries by an average of \texteuro2160 per year but it increased the gap in the rest of the industries by an average of \texteuro780 per year.

\section{Discussion and Conclusion}

This work focused on removing gender bias from word embeddings of vacancy texts and resumes with the goal of creating debiased job recommendations. It showed that gender can be predicted extremely well from anonymised resume embeddings and that naive resume-to-job recommendations based on these embeddings can perpetuate the ``wage gap'' that exists between women and men. Adversarial debiasing improved statistical parity for industry classification based on resume and eliminated the female/male salary difference in job recommendations. This suggests that adversarial debiasing can help make fairer recommendations in realistic scenarios.

Our results indicate that anonymisation alone is not enough to remove indirect information about the gender of the job seeker.
Namely, from our $900k$ anonymised resumes, gender could be predicted with an AUC of $0.94$. This exceeds similar results that have been shown in a smaller data set (AUC=$0.81$)~\cite{parasurama2022gendered}. This is a common problem in fairness-aware machine learning, where removal of directly sensitive information is undermined by correlated features that allow the sensitive information to be inferred~\cite{mehrabi2021survey}.

The difficultly of removing gender bias from language was further illustrated by our data augmentation attempt to substitute a selection of gendered words by neutral words before word2vec training. The resulting embeddings did not effect much change in any of our tests. Previous work on word substitution data augmentation has been shown effective~\cite{hall-maudslay-etal-2019-name,pruksachatkun2021does}, so it may be that our results are limited by the quantity and/or selection of our word substitution pairs (Table \ref{table:naive_method_words}), which were taken from~\cite{WEAT}. While it is possible to improve upon our substitution pairs, creating a complete list of gendered words as used in vacancies and resumes is challenging if not unfeasible, especially in multiple languages.

In contrast, the adversarial approach improved both statistical parity and the wage gap in our data. Using the adversarial representations, prediction of gender dropped from an AUC of $0.94$ to $0.82$ while performance of industry group prediction, in terms of accuracy, dropped only minimally (Table~\ref{tab:fairness_performance}). However, the true positive rate was decreased, indicating that performance was affected. These results are linked and can be adapted by changing the $\alpha$ and $\beta$ parameters in Equation~\ref{eq:1}: more gender-neutral embeddings will likely lead to improved statistical parity but decreased industry prediction performance. Since statistical parity balances for equal true positive rate, the false positive and negative rates are likely to be affected.

Our analysis reveals that ignoring the presence of bias in recruitment texts, that are used to match resumes and vacancies, could lead to severe unwanted discriminatory behaviour. 

The original representations produced a wage gap of \texteuro1680 per year between the female group and the male group.

The adversarial representations eliminated this wage gap to a statistically insignificant difference.

This result is especially important, because it shows that the adversarial representations did not just perform better on selected in-vitro metrics, but also improved fairness in a real application. This suggests that the adversarial representations do not remove bias only ``cosmetically''~\cite{gonen-goldberg-2019-lipstick}, but instead are effective for improving fairness in job recommendation. The adversarial method increased the mean salary for both the female group and the male group, with a higher increase for the female group to balance the gap. This is a positive outcome as the method did not sacrifice the salaries of one of the groups in order to reduce the wage gap.

This work was limited by several factors. Firstly, while the fairness of job recommendations was assessed, due to unavailability of data, the quality of recommendations could not be evaluated. This was mitigated by performing a related classification task: predicting which industry groups a job seeker is interested in based on resume. The accuracy of $0.89$ on this task suggests that salient information relevant to job placement has been preserved. However, since the true positive rate was impacted, it seems likely that the recall of the recommendations would be impacted too. Secondly, the recommender system to suggest jobs based on representation distances was relatively simple; if job-to-resume association data were available, a more complex solution might be preferable. Thirdly, because gender was inferred for this research, it was not possible to include non-binary gender identities~\cite{richards2016non}. Since this group is vulnerable to employment discrimination~\cite{harrison2012gender,doi:10.1080/23311886.2016.1236511,fogarty2018gender}, it should not be overlooked and more research here is needed. Fourthly, results reported in this research use only word2vec document embeddings; other types of embeddings are not considered. Lastly, training of the models was performed using a fixed split instead of cross-validation, which was infeasible due to time and costs. However, the results are likely to be representative given the large size of the data set.

The strengths of our work include the application of adversarial debiasing for fairness-aware machine learning on real and large industry data. While adversarial debiasing for fairness is not novel~\cite{edwards2015censoring,wu2021fairness,liu2022dual,zhang2018mitigating,sattigeri2019fairness}, applications generally extend to publicly available benchmark data sets that make it difficult to assess its applicability to real-world recommendation systems. Our work is one of the first to show the results of adversarial fairness in a real, industrial-scale system. In addition, this research obtained an acceptable trade-off between fairness and performance for a complex multi-label classification task. Finally, this work showed that the adversarial approach eliminated the female/male wage gap in our job recommendations, even though it was not trained for this task.

In conclusion, this work identified gender bias in word representations and salary associations based on recruitment industry texts and successfully applied adversarial debiasing to combat gender bias in job recommendation. With adversarial representations, the mean female/male wage gap was no longer statistically significant due to being reduced by 89\% from \texteuro1680 to \texteuro180 annually. Our results show that adversarial debiasing of word representations can increase real-world fairness of recommendation systems and thus may contribute to creating fairness-aware machine learning systems.

\bibliography{main.bib}

\appendix

\section{Industry and resumes distributions}
\label{app:resumes}
Table \ref{tab:resumes_table_1} shows the distribution of the samples in each industry group over the whole data set. In more than half of the industries the resumes from the female group are under-represented. The ``Technology'' industry has the fewest samples from the female group. The are industries where the resumes from the female group are over-represented, such as the ``Administration/Secretarial'' and the ``Customer service/Call centre/Front office'' industries, where the resumes from the male group are more than three times less present.

\begin{table*}[!htbp]
\small
\caption{Distribution of samples over each industry group. Counts and percentages per industry group do not sum to the expected totals, because job seekers were free to select multiple groups. ``F-M Ratio'' represent the ratio between the number of females within an industry group and the number of males.}
\begin{tabular}{L c c c c}
\toprule
    & \textbf{Male} & \textbf{Female} & \textbf{Total} & \textbf{F-M Ratio}\\
Overall  &  467173 & 437403  &  904576  & 0.93   \\
\midrule
Administration/Secretarial & 45293 \hspace*{0.3cm} (9\%) & 167585 \hspace*{0.1cm} (38\%) & 212878 \hspace*{0.1cm} (23\%) & 3.7\\
Automation/Internet & 49527 \hspace*{0.15cm} (10\%) & 8547 \hspace*{0.50cm} (1\%) & 58074 \hspace*{0.4cm} (6\%) & 0.17 \\
Policy/Executive & 40086  \hspace*{0.3cm} (8\%) & 33541 \hspace*{0.4cm} (7\%) & 73627 \hspace*{0.4cm} (8\%)  & 0.83\\
Security/Defence/Police & 23134 \hspace*{0.3cm} (4\%) & 8821  \hspace*{0.5cm} (2\%) & 31955 \hspace*{0.4cm} (3\%) & 0.38\\
Commercial/Sales & 92801 \hspace*{0.2cm} (19\%) & 66461 \hspace*{0.2cm} (15\%) & 159262 \hspace*{0.1cm} (17\%) & 0.71\\ 
Consultancy/Advice & 69914 \hspace*{0.2cm} (14\%) & 42245 \hspace*{0.3cm} (9\%) & 112159 \hspace*{0.1cm} (12\%) & 0.6 \\
Design/Creative/Journalism  & 19279 \hspace*{0.3cm} (4\%) & 24839 \hspace*{0.3cm} (5\%) & 44118  \hspace*{0.5cm}(4\%) & 1.28\\
Management & 67412 \hspace*{0.3cm} (14\%) & 32153 \hspace*{0.3cm} (7\%) & 99565 \hspace*{0.3cm} (11\%) & 0.48\\
Financial/Accounting & 34233 \hspace*{0.3cm} (7\%) & 25523 \hspace*{0.3cm} (5\%) & 59756 \hspace*{0.4cm} (6\%) & 0.74\\
Financial services & 34342 \hspace*{0.3cm} (7\%) & 29882 \hspace*{0.3cm} (6\%) & 64224 \hspace*{0.4cm} (7\%) & 0.87\\ 
Catering/Retail & 44647 \hspace*{0.3cm} (9\%) & 60588  \hspace*{0.3cm} (13\%) & 105235 \hspace*{0.1cm} (11\%) & 1.35\\
HR/Training & 26852 \hspace*{0.3cm} (5\%) & 53679 \hspace*{0.2cm} (12\%) & 80531 \hspace*{0.4cm} (8\%) & 1.99\\
Procurement/Logistics/Transport & 99429 \hspace*{0.2cm}  (21\%) & 29677 \hspace*{0.3cm}  (6\%) & 129106 \hspace*{0.1cm}  (14\%) & 0.29 \\
Legal & 8638 \hspace*{0.4cm}  (1\%) & 18488 \hspace*{0.3cm}  (4\%) & 27126 \hspace*{0.3cm}  (2\%) & 2.14\\
Customer service/Call centre/Front office & 20000 \hspace*{0.3cm}  (4\%) & 71090 \hspace*{0.3cm}  (16\%) & 91090 \hspace*{0.3cm} (10\%) & 3.55 \\ 
Marketing/PR/Communications & 46832 \hspace*{0.3cm}  (10\%) & 58598  \hspace*{0.3cm}  (13\%) & 105430  \hspace*{0.1cm} (11\%) & 1.25 \\
Medical/Healthcare & 24018 \hspace*{0.4cm} (5\%) & 85414  \hspace*{0.3cm} (19\%) & 109432 \hspace*{0.1cm} (12\%) & 1.25\\
Education/Research/Science & 38430 \hspace*{0.4cm} (8\%) & 66318 \hspace*{0.3cm} (15\%)& 104748 \hspace*{0.1cm} (11\%) & 1.72\\
Other & 86749 \hspace*{0.3cm} (18\%) & 82728 \hspace*{0.3cm}  (18\%) & 169477 \hspace*{0.1cm} (18\%) & 0.95\\
Production/Operational & 77790 \hspace*{0.4cm} (5\%) & 25452 \hspace*{0.3cm} (25\%) & 103242 \hspace*{0.1cm}(11\%) & 0.32\\
Technology & 102798 \hspace*{0.2cm} (22\%) & 9097 \hspace*{0.6cm} (2\%)& 111895 \hspace*{0.1cm} (12\%) & 0.08 \\
\bottomrule
\end{tabular}
\smallskip
\label{tab:resumes_table_1}
\end{table*}

\section{Word-substitution debiasing method}
\label{app:words}
Table \ref{table:naive_method_words} shows the substitutions of the gendered words with the neutral words for both English and Dutch.

\begin{table*}[!htbp]
\caption{Substitutions of gendered words with neutral words used in the word-substitution debiasing method in both English (top) and Dutch (bottom).}
\centering
\begin{tabular}{c c c} 
\toprule
Male Word & Female Word & Neutral Word   \\ [0.5ex] 
 \midrule
 he & she & they \\ 
 his & hers & theirs \\ 
 himself & herself & themselves \\ 
 male & female & person \\ 
 boy & girl & person \\ 
 man & woman & person \\ 
 \midrule
  hij & zij/ze & u \\ 
 zijn & haar & uw \\ 
 hijzelf & zijzelf & uzelf \\ 
 jongen & meisje & persoon \\ 
  man & vrouw & persoon \\
\bottomrule
\end{tabular}
\label{table:naive_method_words}
\end{table*}

\end{document}